\documentclass[a4paper]{svproc}
\pdfoutput=1
\usepackage{url}

\usepackage[font=small,labelfont=bf]{caption}
\usepackage{graphicx}
\usepackage{subcaption}
\usepackage{pgfplots}
\usepackage{tikz}
\usepackage{tikz-3dplot}
\usepackage{hyperref}
\usepackage{amsmath}
\usepackage{cite}
\usepackage{bm}
\usepackage{xcolor}
\usepackage{wrapfig}
\usepackage{cleveref}
\usepackage{nicefrac}
\usepackage{siunitx}
\usepackage[nolist,nohyperlinks]{acronym}
\begin{acronym}
\acro{MAV}{Micro Aerial Vehicle}
\acro{DOF}{Degrees of freedom}
\newacroindefinite{MAV}{an}{a}
\acro{OMAV}{Omnidirectional Micro Aerial Vehicle}
\newacroindefinite{OMAV}{an}{an}
\acro{UAV}{Uncrewed Aerial Vehicle}
\acro{CoG}{Center of Gravity}
\acro{IMU}{Inertial Measurement Unit}
\newacroindefinite{IMU}{an}{an}
\acro{VTOL}{vertical take-off and landing}
\end{acronym}
\begin{document}
\mainmatter              % start of a contribution
\title{Soliro - a hybrid dynamic tilt-wing aerial manipulator with minimal actuators}
\titlerunning{Soliro - a hybrid dynamic aerial manipulator}
\author{Michael Pantic \and Elias Hampp \and Ramon Flammer \and  Weixuan Zhang \and Thomas Stastny \and Lionel Ott \and  Roland Siegwart}
\authorrunning{Michael Pantic et al.} % abbreviated author list (for running head)
\tocauthor{Michael Pantic, Elias Hampp, Ramon Flammer, Weixuan Zhang, Thomas Stastny, Lionel Ott, Roland Siegwart}
\institute{Autonomous Systems Lab, ETH Z\"{u}rich, Switzerland}
\maketitle              % typeset the title of the contribution
\begin{abstract}
The ability to enter in contact with and manipulate physical objects with a flying robot enables many novel applications, such as contact inspection, painting, drilling, and sample collection. Generally, these aerial robots need more degrees of freedom than a standard quadrotor. While there is active research of over-actuated, omnidirectional MAVs and aerial manipulators as well as VTOL and hybrid platforms, the two concepts have not been combined. We address the problem of conceptualization, characterization, control, and testing of a 5DOF rotary-/fixed-wing hybrid, tilt-rotor, split tilt-wing, nearly omnidirectional aerial robot. We present an elegant solution with a minimal set of actuators and that does not need any classical control surfaces or flaps. The concept is validated in a wind tunnel study and in multiple flights with forward and backward transitions. Fixed-wing flight speeds up to 10 m/s were reached, with a power reduction of 30\% as compared to rotary wing flight.
\end{abstract}
\section{Introduction}
Aerial robots capable of manipulation are a useful tool for work at height or in otherwise inaccessible locations. Versatile and stable aerial manipulation is impractical to perform with underactuated, regular \acp{MAV}, due to their attitude and velocity coupling. Omni-directional aerial robots overcome this limitation for 5 and 6 \ac{DOF} flight, however are subject to the same efficiency and range limitations as all other rotary-wing \acp{MAV}. While hybrid, \ac{VTOL} systems address these limitations for regular \acp{MAV}, no equivalent concept has been shown for (nearly) omni-directional systems. In this paper, we present a novel tilt-wing aerial robot named Soliro, capable of both range-efficient forward flight and nearly-omnidirectional (5 \ac{DOF}) flight without additional actuators compared to its rotary-wing version.
\\
In this work, we address the problem of conceptualization, characterisation, control, and testing of a hybrid, tilt-rotor\footnote{Note the different use of the term "tilt-rotor" in the two fields - in \ac{VTOL} literature they are only used for transitioning, whereas in aerial manipulation tilt-rotors are used for generic thrust vectoring.}, tilt-wing, aerial robot. While both flight modes, over-actuated rotary-wing flight and forward fixed-wing flight, differ significantly in actuation and dynamics, we present an elegant solution with a minimal set of actuators that are all used in both flight modes, and allow seamless transition between the flight modes.
We contribute the wing design and aerodynamic model identification, unifying control strategy, and experimental verification for a hybrid thrust-vectoring aerial manipulator. Compared to classical tilt-wing hybrids, the presented approach uses the tilt mechanism as a full-speed actuator rather than a slow transition between modes~--~effectively blending fixed-wing and overactuated rotary-wing flight. In this work, we concentrate on the fixed-wing and transitioning aspects of Soliro.
\begin{figure}
\begin{subfigure}{.5\textwidth}
\centering
   \includegraphics[width=\linewidth]{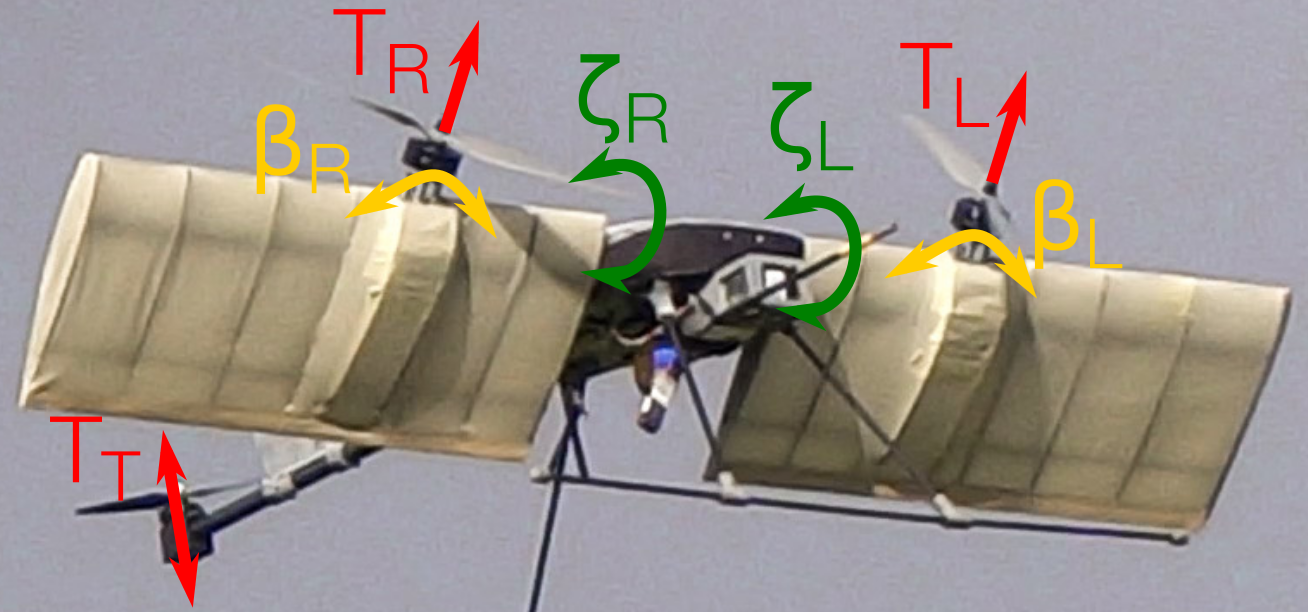}
    \caption{Actuation and system overview. Green $\zeta_{L,R}$ are the arm/wing rotation axes, red $T_{L,R,T}$ are the propeller thrust, and yellow $\beta_{L,R}$ the rotor-tilt servos for side-ways translation. $\beta_{L,R}$ is not used for fixed-wing flight.}
    %\vspace{-20pt}
    \label{fig:soliro}
\end{subfigure}\hfill 
\begin{subfigure}{.45\textwidth}
  \centering
  \includegraphics[width=\linewidth]{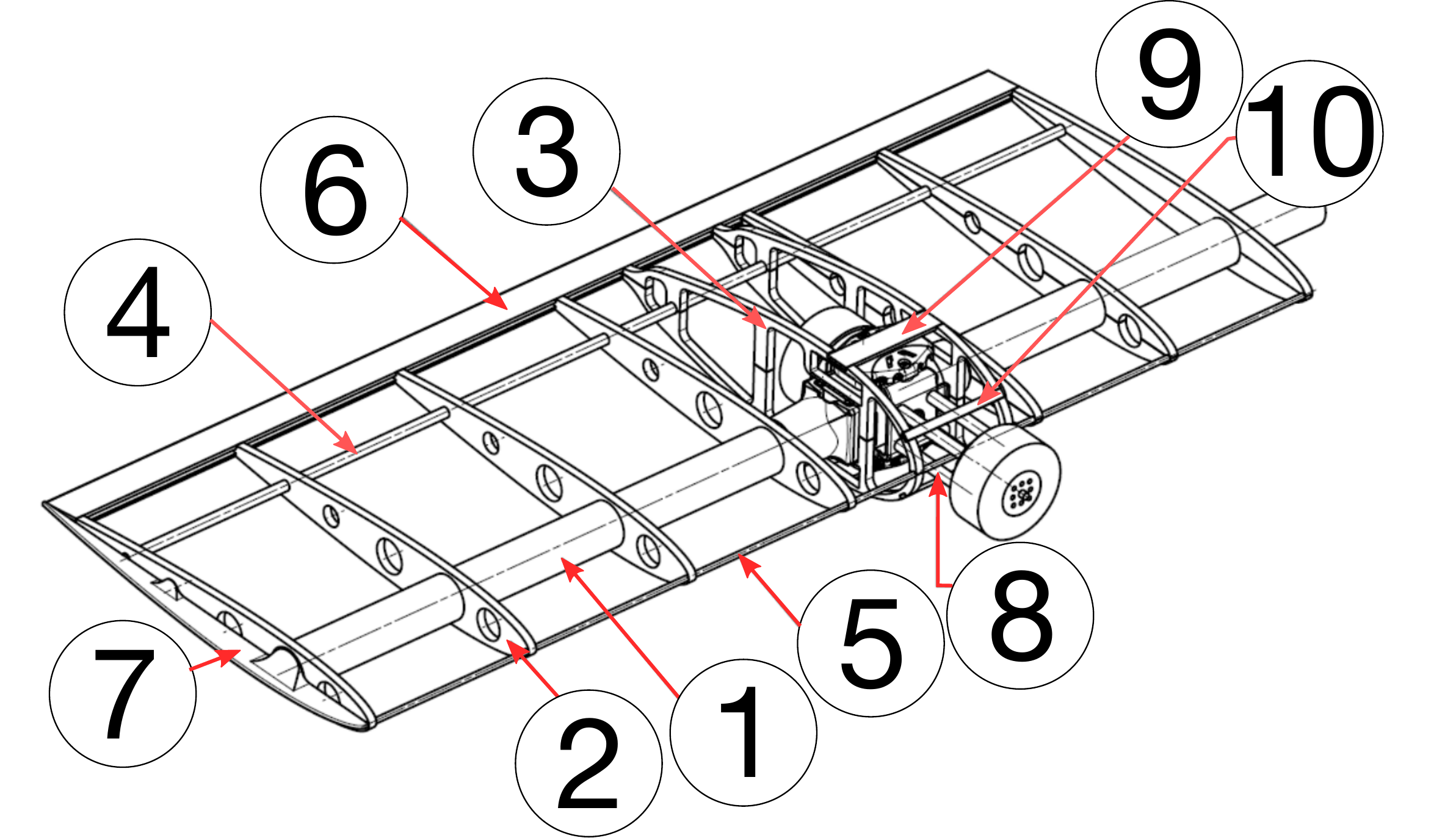}
  \caption{Wing CAD model. (1) Main spar, (2) NACA 0012 rib, (3) NACA 0029
rib, (4) Supporting spar, (5) Leading edge, (6) Trailing edge, (7) Wingtip, (8) Motor
spacer, (9) \& (10) Skin guidance.}
  \label{fig:wing}
\end{subfigure}%
\end{figure}
\subsection*{Related Work}
Tilt-wing, tilt-rotor, and transitional quad-copter systems are common concepts used in \ac{VTOL} capable hybrid aerial vehicles \cite{kamal2018design}. Typical tilt-wing systems \cite{9444145} use a slow actuator to change the angle of incidence between the wing and fuselage for transition, and classical flight controls in fixed-wing mode. In most of the \ac{VTOL} literature, the focus lies in alleviating the need for runways and adding hover abilities to fixed-wing vehicles \cite{7158828}, or to extend the flight range of rotary-wing \acp{MAV}.
Simultaneously, rotary-wing aerial manipulation using thrust vectoring with tilt rotors and over-actuated systems have been an active research topic in the last years. Most of these systems either use tilt-able rotors \cite{9295362, zheng2020tiltdrone} or fixed-mounted rotors in non-orthogonal directions \cite{8401328,brescianini2018omni} to achieve the necessary force and torque envelopes.
However, the two research streams of \ac{VTOL} hybrids and omni-directional \ac{MAV} have not been combined so far.
\section{Technical Approach}
We use the previously existing 5-DOF tilt-rotor Voliro Tricopter as the base for hardware development \cite{9495951}. It consists of two arms that provide 2D thrust vectoring each, and a reversible tail rotor. Each arm can be turned independently $(\zeta_{L,R})$, and each motor can be tilted sideways $(\beta_{L,R})$. This enables free movement in 5 DOF (3D position, pitch, and yaw). \Cref{fig:soliro} shows the platform and its degrees of freedom. To convert this platform into a fixed-wing flight capable robot, we developed a novel wing and actuation design, together with a completely new controller capable of handling all flight states. In this chapter, we present the wing design, system identification, and the control approach used.
\subsection{Wing design}
Re-using the existing arm-tilting servos $(\zeta_{L,R})$ for wing tilt was a major design goal. The wing has to accommodate the existing propulsion system and mechanics, and has to be symmetrical to avoid undesired longitudinal lift in rotary-wing flight. Its size should be small enough to not hinder manipulation. We investigated a range of symmetric NACA profiles\cite{jacobs1937airfoil} in an optimization study, where for every combination of wingspan and chord length the optimal incident angles for range and endurance were determined, and subsequently the required flight velocity and rotor thrust was calculated. Finally, a NACA0012 wing with wingspan 1.5 m, chord length 0.26 m and aspect ratio 5.77 was chosen, with NACA0029 rotor nacelles. 
\Cref{fig:wing} shows a CAD rendering of the wing, which is manufactured using the classic rag-and-tube method. We used carbon fiber spars, with wooden laser-cut ribs and Oratex fabric covering.

\subsection{Actuation design}
Careful analysis of a many design variants revealed that all body axes (yaw, pitch, roll) needed for fixed-wing flight can be controlled without additional actuators as compared to the pre-existing Tri-copter, completely avoiding the need for traditional flight control surfaces. We control yaw by differential thrust, and roll by differential wing tilt. Augmented pitch control is achieved through an active controller that uses the tail rotor with a reversible propeller as its actuator. 
\subsection{Model identification}
We model the complete influence of the wings as a grey-box model that is calibrated using wind tunnel data in a non-linear least-squares fit.
The overall model roughly follows the airfoil model from~\cite{8703808}, where, depending on the angle-of-attack, a flat-plate model is used in stall conditions and a regular airfoil model is used for non-stall conditions. As the wing consists of two different profiles, we use a segment-wise mixture of the corresponding models.
Compared to a conventional airplane, Soliro has a much higher propeller diameter to wingspan ratio. Therefore, a large part of the wing lies within the down wash of the propeller, which affects airspeed over that segment of the wing. After the first outdoor flights, we augmented the model to consider these effects. We calculate the effective corrected air velocity component over the wing, $v_{z,wing,tot} = \sqrt{v_{z,wing}^2 + \frac{4 T}{\pi \rho D^2}}$, where $v_{z,wing}$ is the regular airspeed component parallel to the chord line, $\rho$ the air density, $T$ the propeller thrust, and $D$ the propeller diameter. The effective angle of attack is then calculated as  $\alpha_{effective} = atan2(v_{x,wing}, v_{z,wing,tot})$, with $v_{x,wing}$ being the perpendicular air speed component to $v_{z,wing}$, and the total corrected air velocity as $\sqrt{v_{z,wing,tot}^2 + v_{x,wing}^2}$. These quantities are then used to calculate the effective lift for the segments affected by prop wash, which lie within $\frac{D}{\sqrt{2}}$ of the propeller axis.
\subsection{Control architecture}
The control architecture of Soliro centers around a unified control allocation scheme, which is valid throughout the complete flight envelope. The unified allocation outputs the setpoints for all actuators (motors thrusts $T_r$, $T_l$, $T_{t}$ and wing tilts $\zeta_r$, $\zeta_l$) based on the desired body torques $\tau_{roll}$, $\tau_{pitch}$, $\tau_{yaw}$, overall thrust $T_{col}$, and overall wing tilt $\chi$. 
\begin{figure}[ht]
    \centering
     \scalebox{1}{%
        \input{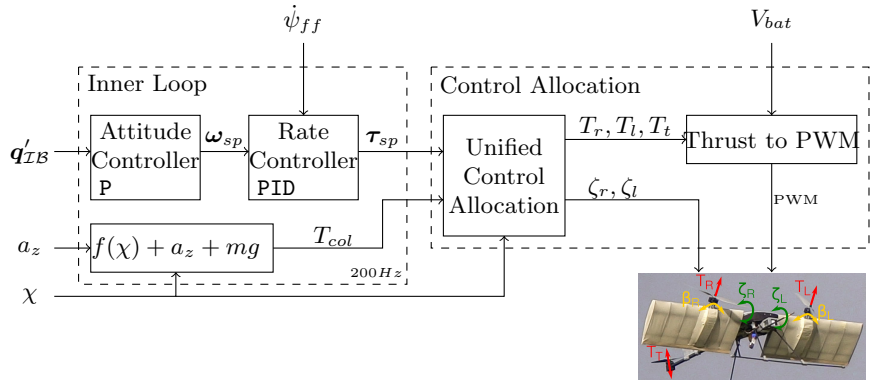}
        }
    \caption{The complete low-level control and allocation model of Soliro. The inputs are the desired wing overall wing tilt $\chi$, vertical acceleration $a_z$, desired attitude $\boldsymbol{q}_{\mathcal{I}\mathcal{B}}'$.}
    \label{fig:inner_loop}
\end{figure}
The high-level controller or the user can select $\chi$ freely to continuously choose between fixed-wing and rotary-wing flight. \Cref{fig:inner_loop} gives an overview of the low-level controller architecture of Soliro.
\subsection{Unified Allocation}
We obtain the unified allocation by modeling the aerodynamic and propulsion forces and torques acting on each body axis as a function of the system state, and then solving for the actuator outputs. 
\begin{figure}[ht]
  \centering
  \scalebox{0.9}{%
        \tdplotsetmaincoords{-80}{200}
\begin{tikzpicture}[>=latex,tdplot_main_coords, scale=1.3, rotate=0]
\tikzstyle{every node}=[font=\large]
\pgfmathsetmacro{\xlen}{1.0}
\pgfmathsetmacro{\ylen}{1.6}
\pgfmathsetmacro{\zlen}{1.0}
\pgfmathsetmacro{\b}{4.0}
\pgfmathsetmacro{\l}{5.0}

\pgfmathsetmacro{\Ttail}{1.0}
\pgfmathsetmacro{\Tright}{3.0}
\pgfmathsetmacro{\Tleft}{3.0}
\pgfmathsetmacro{\TauAero}{1.5}

\pgfmathsetmacro{\costhirty}{0.86602540378}
\pgfmathsetmacro{\chides}{30}
\pgfmathsetmacro{\epsi}{10}

\pgfmathsetmacro{\xoff}{-2}
\pgfmathsetmacro{\yoff}{1}
\pgfmathsetmacro{\zoff}{1}

\definecolor{red}{HTML}{a66259}
\definecolor{blue}{HTML}{8bb4b4}
\definecolor{yellow}{HTML}{f5e499}

\draw[line width = 2,->] (0,0,0) -- (\xlen,0,0) node[above]{$\boldsymbol{x}$};
\draw[line width = 2,->] (0,0,0) -- (0,\ylen,0) node[below]{$\boldsymbol{y}$};
\draw[line width = 2,->] (0,0,0) -- (0,0,\zlen) node[anchor=north east]{$\boldsymbol{z}$};
\draw[line width = 2,->, red] (\xoff,\yoff,\zoff) -- (\xoff + \xlen,\yoff,\zoff) node[above]{$\boldsymbol{\tau}_{roll}$};
\draw[line width = 2,->, red] (\xoff,\yoff,\zoff) -- (\xoff,\yoff + \ylen, \zoff) node[below left]{$\boldsymbol{\tau}_{pitch}$};
\draw[line width = 2,->, red] (\xoff,\yoff,\zoff) -- (\xoff, \yoff, \zoff + \zlen) node[below]{$\boldsymbol{\tau}_{yaw}$};
\draw[line width = 1] (0,0,0) -- (0,\b, 0);
\draw[line width = 1] (0,0,0) -- (0,-\b, 0);
\draw[line width = 1] (0,0,0) -- (-\l, 0, 0);
\draw  (0,0,0) node[above] {$\mathcal{S}$};
\draw  (\xoff,\yoff,\zoff) node[above left] {$\mathcal{B}$};

\draw[dashed] (0,\b, 0.0) -- ({\Tright*sin(\chides)}, \b, {-\Tright*cos(\chides)});
\draw[dashed] (0,-\b, 0.0) -- ({\Tright*sin(\chides)}, -\b, {-\Tright*cos(\chides)});

\draw[line width = 2, ->, yellow] (0,0, 0.0) -- ({\TauAero*sin(\chides)}, 0, {-\TauAero*cos(\chides)}) node[below left]{$\boldsymbol{\tau}_{aero}$};

\draw[dashed] (0,-\b, 0.0) -- ({\Tright*sin(\chides)}, -\b, 0);
\draw[dashed] (0,\b, 0.0) -- ({\Tleft*sin(\chides)}, \b, 0);

\draw[thick, ->, blue] (0,\b, 0.0) -- ({\Tright*sin(\chides-\epsi)}, \b, {-\Tright*cos(\chides-\epsi)})node[above right]{$\boldsymbol{T}_l$};
\draw[thick, ->, blue] (0,-\b, 0.0) -- ({\Tleft*sin(\chides+\epsi)}, -\b, {-\Tleft*cos(\chides+\epsi)})node[above left]{$\boldsymbol{T}_r$};

\tdplotsetthetaplanecoords{0}
\tdplotdrawarc[tdplot_rotated_coords, ->, red]{(0, 0, \b)}{1.2}{90}{180-\chides}{below right}{$\chi$}
\tdplotdrawarc[tdplot_rotated_coords, ->, red]{(0, 0, -\b)}{1.2}{90}{180-\chides}{below right}{$\chi$}
\tdplotdrawarc[tdplot_rotated_coords, ->, blue]{(0, 0, \b)}{1.5}{90}{180-\chides + \epsi}{left}{$\zeta_l$}
\tdplotdrawarc[tdplot_rotated_coords, ->, blue]{(0, 0, -\b)}{1.5}{90}{180-\chides - \epsi}{left}{$\zeta_r$}
\tdplotdrawarc[tdplot_rotated_coords, <-]{(0, 0, -\b)}{1.8}{180-\chides - \epsi}{180-\chides}{above left}{$\epsilon$}
\tdplotdrawarc[tdplot_rotated_coords, <-]{(0, 0, \b)}{1.8}{180-\chides + \epsi}{180-\chides}{above left}{$\epsilon$}

% CoG 
\draw[dashed] (0.0, \yoff, 0.0) -- (\xoff, \yoff, 0);
\draw[dashed] (\xoff, \yoff, 0.0) -- (\xoff, 0.0, 0);
\draw[dashed] (\xoff, \yoff, 0.0) -- (\xoff, \yoff, \zoff);
\draw[thick](\xoff, \yoff, \zoff) node[] {\textbullet};
\draw[thick](\xoff, \yoff, \zoff) node[right] {$CoG$};
\draw [decorate, decoration = {brace, mirror}] (0,\yoff,0) --  (\xoff, \yoff, 0);
\draw  ({\xoff/2} , \yoff, 0) node[below] {$x_{off}$};
\draw [decorate, decoration = {brace}] (\xoff,0,0) --  (\xoff, \yoff, 0);
\draw  ({\xoff} , {\yoff/2}, 0) node[below right] {$y_{off}$};
\draw [decorate, decoration = {brace, mirror}] (\xoff,\yoff,0) --  (\xoff, \yoff, \zoff);
\draw  ({\xoff} , {\yoff}, {\zoff/2}) node[left] {$z_{off}$};

\draw [decorate, decoration = {brace}] (0,0,0) --  (0,-\b,0);
\draw  (0,{-\b/2},0) node[above left] {$b$};
\draw [decorate, decoration = {brace}] (0,0,0) --  (-\l,0,0);
\draw  ({-\l/2},0,0) node[above] {$l$};

\draw[thick, ->, blue] (-\l, 0, 0) -- (-\l, 0, -\Ttail) node[above right]{$\boldsymbol{T}_{t}$};
\draw[thick,->, red] (\xoff,\yoff,\zoff) -- (\xoff, \yoff, \zoff - 3) node[above]{$\boldsymbol{T}_{collective}$};

\end{tikzpicture}
        }
  \caption{The essential forces and torques acting on Soliro. Blue indicates the output of the control allocation, i.e. the forces and angles that can directly be controlled by the actuators. Red indicates the input to the control allocation. The yellow arrow is the aerodynamic torque created by differential wing deflection}
  \label{fig:control_allocation}
\end{figure}
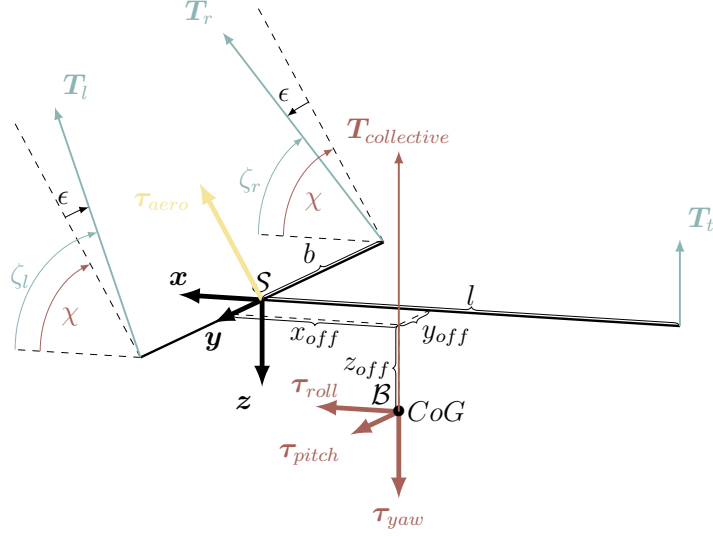
To obtain a simple, closed-form solution, we use small angle approximations of the differential wing deflections $\epsilon$ and decouple the pitch axis from the rest. We consider the thrust of the tail rotor only for the torque around the $y$-axis, but not for overall body thrust. The made approximations and simplifications are appropriate, as the differential wing deflections are generally very small and the tail rotor's location is far away from the \ac{CoG} and only thrusts intermittently. \Cref{fig:control_allocation} shows the coordinate frames and quantities used. 
%\vspace{-3pt}
Given a set of desired body torques ($\tau_{roll}, \tau_{yaw},\tau_{pitch}$), overall wing tilt $\chi$, and desired total thrust $T_{col}$, we obtain the left and right thrust $T_r,T_l$, and the differential wing deflection $\epsilon$ as
\begin{align*}
   T_r &= \frac{(T_{col}~b + g~m~y_{offset} - \tau_{roll})~\sin{\left(\chi \right)} - \tau_{yaw} \cos{\left(\chi \right)}}{2 b} \\
   T_l &= \frac{(T_{col}~b - g~m~y_{offset} + \tau_{roll})~\sin{\left(\chi \right)} + \tau_{yaw} \cos{\left(\chi \right)}}{2 b} \\
   \epsilon &= \frac{(g~m~y_{offset}- \tau_{roll})~\cos{\left(\chi \right)} + \tau_{yaw} \sin{\left(\chi \right)}}{T_{col}~b + \tau_{aero}}.
\end{align*}
Based on the obtained thrusts and differential wing deflection and desired pitch torque $\tau_{pitch}$ , the tail rotor thrust calculates as
\begin{multline*}
    T_t = - \frac{T_{r} x_{offset} \sin{\left(\chi + \epsilon \right)}}{l + x_{offset}} + \frac{T_{r} z_{offset} \cos{\left(\chi + \epsilon \right)}}{l + x_{offset}} \\- \frac{T_{l} x_{offset} \sin{\left(\chi - \epsilon \right)}}{l + x_{offset}} + \frac{T_{l} z_{offset} \cos{\left(\chi - \epsilon \right)}}{l + x_{offset}} - \frac{\tau_{pitch}}{l + x_{offset}}.
\end{multline*}
The system is then commanded using the thrusts $T_r, T_l, T_t$, and the total wing angles $ \zeta_{r} = \chi + \epsilon, \zeta_{l} = \chi - \epsilon$. The \ac{CoG} offsets in the equations above help to trim the vehicle.
We use $\tau_{aero}$ to describe the control authority of the wing deflection for roll - which depends on the airspeed and is empirically modelled as a quadratic approximation. We used the function $\tau_{aero} =  \frac{200}{(\chi_{min}-0.5\pi)^{2}} (\chi - 0.5\pi)^{2}$, where $\chi_{min}$ is the minimal overall wing deflection of about \SI{10}{deg}. Due to the direct correlation between airspeed for level flight and overall wing deflection $\chi$, we chose to model $\tau_{aero}$ directly based on the wing deflection and circumvent the need for a high-accuracy airspeed sensor. However, depending on the sensor setup, $\tau_{aero}$ could be modeled based on pitot tube readings.
\subsection{Low-level attitude and thrust control}
The inner loop of the controller consists of a cascaded controller with an attitude P controller and a rate PID controller, shown in Figure \ref{fig:inner_loop}. The attitude controller converts the difference of desired attitude $\boldsymbol{q}_{\mathcal{I}\mathcal{B}}'$ and current attitude $\boldsymbol{q}_{\mathcal{I}\mathcal{B}}$ into angular rates, which then are fed into the rate controller. The attitude control law is based on~\cite{brescianini2013nonlinear,px4_autopilot}, and formalized as:
\begin{align*}
        \boldsymbol{q}_{error}  &= \boldsymbol{q}_{\mathcal{B}\mathcal{B}'}= \boldsymbol{q}_{\mathcal{I}\mathcal{B}}^{-1} \otimes \boldsymbol{q}_{\mathcal{I}\mathcal{B}'}. \\
        \boldsymbol{\omega}_{sp} &= \boldsymbol{k}_P\check{\boldsymbol{q}}_{error}\quad \text{with } \boldsymbol{k}_P = \begin{pmatrix}k_{P,x} & 0 & 0\\0& k_{P,y}& 0 \\ 0& 0& k_{P,z} \end{pmatrix},
\end{align*} with error quaternion $\boldsymbol{q}_{error}$ respectively its imaginary part $\check{\boldsymbol{q}}_{error}$, gain matrix $\boldsymbol{k}_P$. The resulting angular rate setpoint $\boldsymbol{\omega}_{sp}$ is then fed into the rate controller to obtain the necessary body torques for the allocation.
The body torques are calculated as
\begin{equation*}
\begin{aligned}
    \boldsymbol{\tau}_{sp} &= \boldsymbol{k}_{P,rate}(\boldsymbol{\omega} - \boldsymbol{\omega}_{sp}) - \boldsymbol{k}_{D,rate}\boldsymbol{\dot\omega} + \boldsymbol{\omega}_{int}\\
    \text{with } \boldsymbol{\omega}_{int} &= \sum^{N}_{i=1}\boldsymbol{k}_{I,rate}(\boldsymbol{\omega}_i - \boldsymbol{\omega}_{i,sp})\delta t.
\end{aligned}
\end{equation*}
$\boldsymbol{k}_{P,rate}$, $\boldsymbol{k}_{I,rate}$, and $\boldsymbol{k}_{D,rate}$ are the inertia-normalized gain matrices and $\delta t$ is the controller time step duration.
In addition to the attitude and rate control, we add a roll-axis trim feed-forward to account for wing manufacturing asymmetries as well as a thrust feed-forward that that takes into account the lift generated by the wings. The roll trim is implemented as a feed-forward torque on the roll axis,  $\boldsymbol{\tau}_{sp,roll} \mathrel{+}=  \boldsymbol{\tau}_{trim,roll} \cos(\chi)$, which is then scaled similarly to $\tau_{aero}$ to account for the asymmetries' impact at different airspeeds.
\\
The collective thrust $T_{col}$ fed into the allocation is calculated by the feed-forward term $T_{ff}$ and the desired acceleration in $z$ as 
\begin{equation*}
T_{col} = \frac{-a_z T_{ff}}{g} + T_{ff} \qquad T_{ff}= \sum{}_{k=0}^{7}(c_{i}~\chi^{k}),  
\end{equation*} 
$T_{ff}$ is a wind tunnel identified, 7\textsuperscript{th} order polynomial that expresses the relationship between overall wing tilt $\chi$ and the thrust reduction/feed-forward for forward flight. Our experimental results have shown that the model is fairly accurate and robust, despite using $\chi$ as a proxy for airspeed.
\\
Compared to a normal airplane, the main motors of Soliro in practice always have a thrust component in $z$, i.e. they never purely point forward. This allows direct control over the altitude of the vehicle, even in fixed-wing mode. Contrastingly, fixed-wing airplanes often use an altitude controller which considers conservation of kinetic and potential energy, e.g. a TECS controller  \cite{bruce1986nasa}.
The advantage of the here propose altitude controller is its validity over the complete flight envelope while the implementation of a TECS controller would be very cumbersome for a vehicle like Soliro.
\section{Experiments}
We verified the design and aerodynamic properties in an extensive wind tunnel characterization. In a series of field tests, we verified the control framework, transition stability, and forward flight efficiency under realistic conditions.
\subsection{Wind tunnel characterization}
We use a wind tunnel with adjustable flow velocity and angle to fully identify and verify the chosen wing design and overall system in terms of lift, drag, stall behaviour, and stability. Importantly, controllability and input sensitivity of the minimal actuator concept needed to be verified, e.g. strength of differential wing tilt and lateral airflow impact on the tail rotor. We use the resulting 26 parameter model of the vehicle and its actuators for realistic simulation and off-line controller tuning.
\begin{figure}
\begin{subfigure}{.48\textwidth}
 \includegraphics[width=\linewidth]{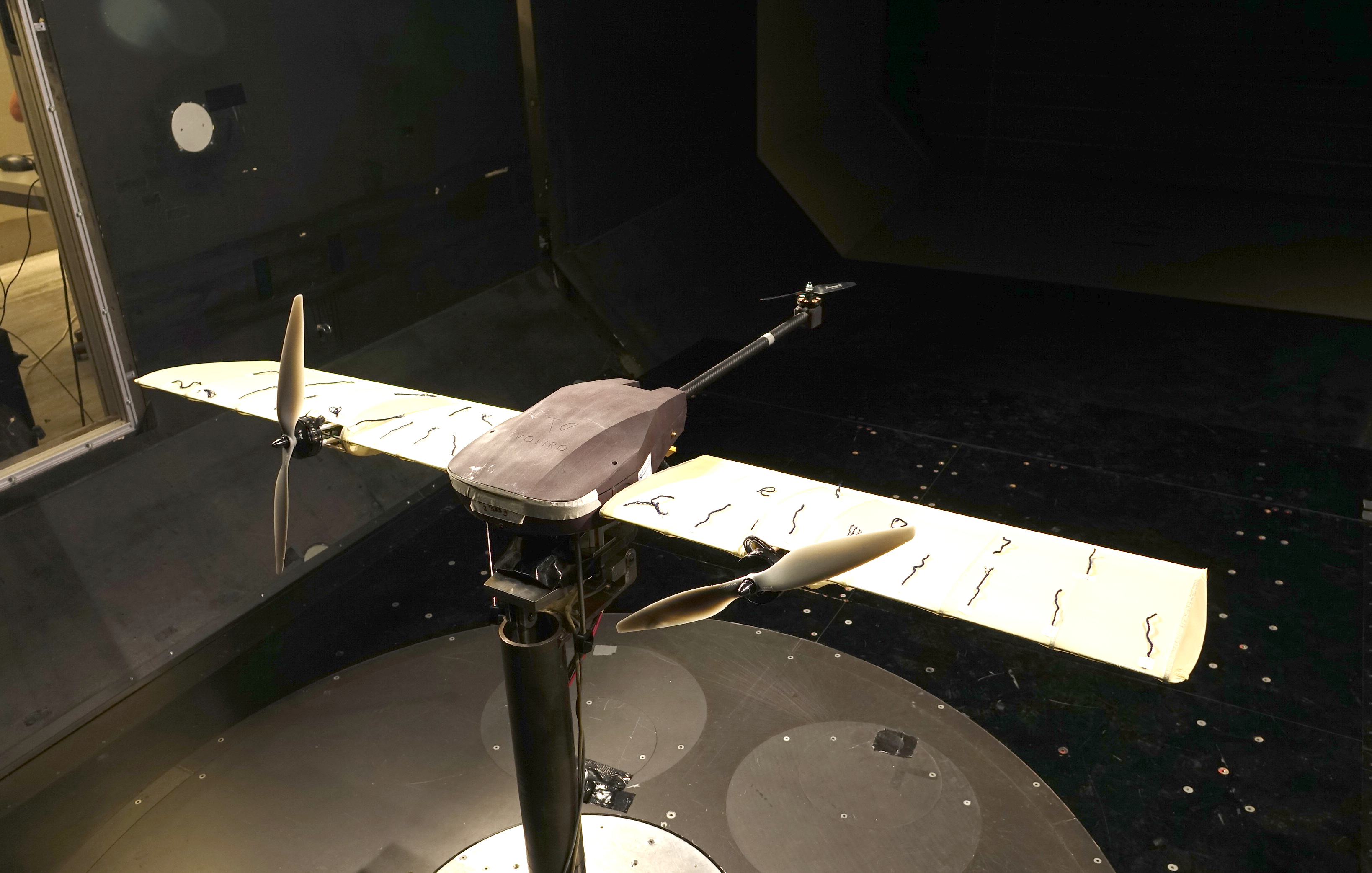}
    \caption{Wind tunnel set-up including yarns on wing to visualize air flow. A 6-DOF Force-Torque sensor for full system identification is mounted below vehicle.}
   \label{fig:windtunnel}
\end{subfigure}\hfill
\begin{subfigure}{.48\textwidth}
  \centering
  \includegraphics[width=\linewidth]{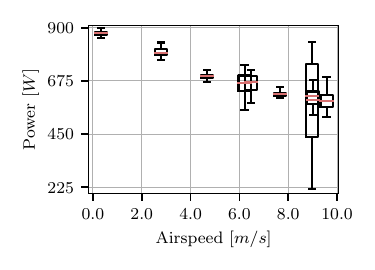}
  \caption{In-flight measured total power statistic for altitude hold at different air speeds during multiple \textbf{outdoor} flights. Airspeed of \SI{0}{\meter\per\second} corresponds to rotary wing hover.}
  \label{fig:airspeed_power}
\end{subfigure}%
\end{figure}
The grey box model is calibrated over a large range of flow speeds and angles in the wind-tunnel. The wing performs well and is able to provide sufficient lift, although the first principle model slightly overestimates the maximum lift. According to the lift/drag curves and a yarn string test, stall onset is very\ gradually. 
To test the maneuverability, we identified sensitivity and resolution for all actuators. Specifically, we conclude that the differential wing tilt enables sufficiently gradual roll torque and stability. We obtain an almost ideal linear fit of \SI{0.45}{\newton\meter} per \SI{1}{\degree} of differential wing tilt, with the smallest stable increment being \SI{0.08}{\degree} differential wing tilt, corresponding \SI{0.038}{\newton\meter} of torque.
Additionally, we used the wind-tunnel experiments to test stall characteristics, maximum servo- and wing loadings, as well as actuator efficiency and discretization.
\subsection{Transition and Flight}
\begin{figure} [h]
\centering
\begin{subfigure}{.48\textwidth}
    \includegraphics[width=\linewidth]{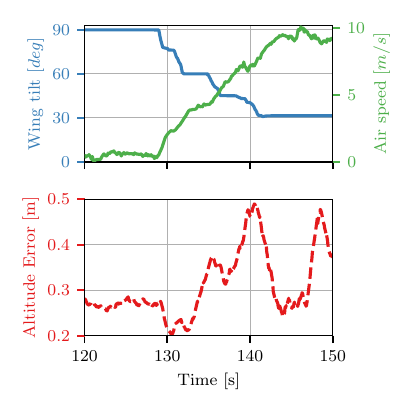}
     \caption{Transition behaviour of the controller. The wing tilt is controlled by the pilot, whereas the airspeed is automatically increased to hold altitude. The approx. 28 cm static altitude error offset is due to weight calibration.}
    \label{fig:transition_behaviour}    
\end{subfigure}\hfill
\begin{subfigure}{.48\textwidth}
    \includegraphics[width=\linewidth]{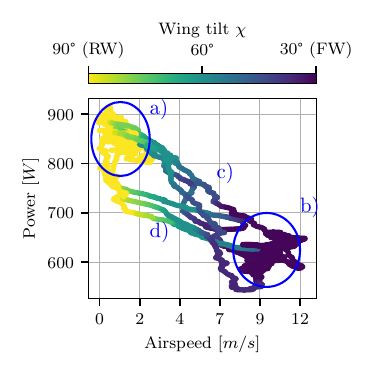}
    \caption{Detailed analysis of power efficiency for three transitions in a single flight. Color indicates overall wing tilt $\chi$. Region a) is pure rotary-wing flight, b) pure fixed-wing flight. Region c) corresponds to the forward transition (RW to FW), and d) to the backward transition.}
    \label{fig:power_detailed}
\end{subfigure}%
\end{figure}

We performed a series of outdoor tests where the system starts in rotary-wing mode and then slowly transitions into fixed-wing flight mode with up to \SI{10}{\meter\per\second} flight speed and back. These experiments are used to verify the control strategy, stability of mode transitions, and fixed-wing mode flight efficiency and models. 
\begin{figure}[ht]
    \centering
    \includegraphics[width=\linewidth]{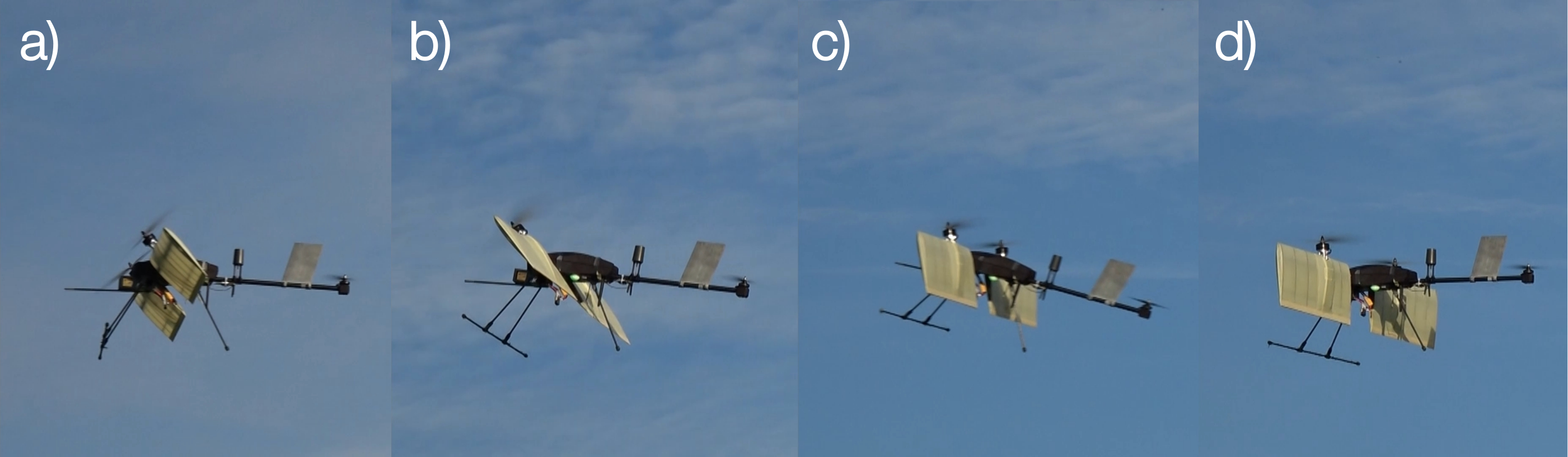}
    \caption{The system performing a reverse transition from fixed-wing (a) to rotary-wing mode (d). Longitudinal stability and flight path are decoupled from body pitch angle, the body can be used  as spoiler to reduce air speed during reverse transition (c).}
    \label{fig:transition}
\end{figure}
The outdoor experiments were performed under realistic conditions, including wind and turbulence. \Cref{fig:transition} shows a transition maneuver in flight. After some tuning and a few tries with bugs and other issues, the system flies smooth and docile and all axes all stable and controllable. \Cref{fig:transition_behaviour} visualizes the detailed controller performance for a forward transition.
\subsection{Efficiency}
\Cref{fig:airspeed_power} shows the power usage in different flight conditions. The system is able to fly steadily at \SI{10}{\meter\per\second} with about a $30\%$ thrust reduction compared to the rotary-wing hovering state. \Cref{fig:power_detailed} visualizes the power consumption, wing-tilt, and air speed in detail for a single flight. Note that there is also rotary-wing forward flight visible in region a). Another interesting observation is that during backward transition energy has to be dissipated, therefore low transient rotary-wing power consumption is observed (below region a)). Additionally, the plot shows that there is a good correspondence between wing-tilt $\chi$ and altitude-hold air speed, which is exploited in the control architecture to omit the need for an airspeed sensor.
\section{Conclusion \& Discussion}
In this paper, we presented a hybrid dynamic tilt-wing aerial manipulator with minimal actuators, including the design, model approach and control. In fixed-wing flight, the system demonstrated a power reduction of roughly $30\%$ over rotary-wing flight. The power-reduction corresponds to a range of about $\SI{10}{\kilo\meter}$ on a single battery - aerodynamic modeling indicates that further optimizing the drag of the platform (landing legs, antennas) would increase this range to up to $\SI{30}{\kilo\meter}$.
A main insight of the experiments is the confirmation that the chosen actuator strategy, without any classical control surfaces, performs well. The conducted wind tunnel verification was of great importance for verification, calibration and tuning prior to first flight.
Despite all possible precautions, we encountered a few crashes. On the very first flight, the pitch stability controller reacted too strongly based on our model calculations and tunings. All other crashes thereafter were related to hardware or remote control issues. 
\\
The actuator design with the augmented pitch stabilization controller and the fast (differential) tilt-wing dynamics creates an extremely large flight envelope, which contains maneuvers impossible with traditional tilt-wing platforms. For example, the wings or the body can be used to rapidly increase drag, effectively becoming a spoiler. Similarly, the active pitch controller can counter any longitudinal static or dynamic instabilities due to weight distribution or external disturbances.
\bibliographystyle{ieeetr}
\bibliography{references}
\end{document}